\documentclass{article}

    \PassOptionsToPackage{numbers, compress}{natbib}

    \usepackage[preprint]{neurips_2022}

\usepackage[utf8]{inputenc} 
\usepackage[T1]{fontenc}    
\usepackage{hyperref}       
\usepackage{url}            
\usepackage{booktabs}       
\usepackage{amsfonts}       
\usepackage{nicefrac}       
\usepackage{microtype}      
\usepackage{xcolor}         

\usepackage{graphicx}
\usepackage{amsmath}
\usepackage{amssymb}
\usepackage{bbm}
\usepackage{bm}
\usepackage{booktabs}
\usepackage{xcolor}
\usepackage{array}
\usepackage[labelfont=bf,font=small,skip=3pt]{caption}
\newcolumntype{C}[1]{>{\centering\arraybackslash}p{#1}}
\usepackage{wrapfig}
\usepackage{subcaption}

\title{The Progression of Transformers from Language to Vision to MOT: A Literature Review on Multi-Object Tracking with Transformers}

\author{%
  Abhi Kamboj \\
  Department of Electrical and Computer Engineering\\
  University of Illinois\\
  Champaign, IL 61820 \\
  \texttt{akamboj2@illinois.edu} \\
}

\begin{document}

\maketitle

\begin{abstract}
The transformer neural network architecture allows for autoregressive sequence-to-sequence modeling through the use of attention layers. It was originally created with the application of machine translation but has revolutionized natural language processing. Recently, transformers have also been applied across a wide variety of pattern recognition tasks, particularly in computer vision. In this literature review, we describe major advances in computer vision utilizing transformers. We then focus specifically on Multi-Object Tracking (MOT) and discuss how transformers are increasingly becoming competitive in state-of-the-art MOT works, yet still lag behind traditional deep learning methods.
\end{abstract}

\section{Introduction}
A vital part of explainable video understanding is being able to detect and track objects in a video. This yields the problem of object tracking. With the recent rise of the transformer architecture in deep learning, one may posit: how can the transformer architecture be leveraged to design effective object-tracking methods?

In this literature review, we describe the progression of the transformer idea from language, through computer vision and into multi-object tracking. We begin introducing transformers in Section~\ref{sec:transformers}. We further discuss how they progressed into the field of computer vision, specifically object detection in Section~\ref{sec:cv}. Finally in  Section~\ref{sec:mot}, we delve into the problem of multi-object tracking. In particular, we highlight how the same concepts that evolved from language transformers to vision transformers are being applied in multi-object tracking methods. Despite this evolution of tranformers and self-attention, the best, highest performing, object tracking methods do not use transformers, indicating they may not be the best tool for multi-object tracking.

\section{Introduction to Transformers}
\label{sec:transformers}
\subsection{B.T.: Before Transformers}
Sequence to sequence modeling is the problem in which one sequence of inputs needs to be translated to a sequence of outputs. A common example of this is in machine translation, i.e. translating from one language to another. For many years, recurrent neural networks (RNNs) performed the best for these types of tasks. RNNs are uniquely able to input and output sequences of varying length by sharing parameters through time, i.e. an output depends on the previous time step's output as well. This method is able to perform time series prediction through updating and using hidden states \cite{elman1990finding}. 

The main issue with RNNs was that back-propagating the gradient through numerous time steps led to diminishing gradient values that negligibly contributed to gradient update steps, making the model difficult to train. Since the error vanishes through time this problem was termed vanishing gradients \cite{hochreiter1998vanishing}. Many iterations of research worked on this issue and improved on the problem. Long short term memory (LSTM) introduced a set of learned gates to input or reset the hidden states and showed to capture long term dependencies better \cite{hochreiter1997long} and Gated Recurrent Units (GRUs) performed on par with LSTMs using significantly less gates and memory \cite{chung2014empirical}. 

For many decades these RNN based auto-regressive models performed state of the art (SOTA) until Vaswani et al. 2017 \cite{vaswani2017attention} introduced transformers. A transformer applies the concept of attention in a novel fashion to a input of sequence. Compared to traditional RNNs, the attention mechanism essentially avoids the vanishing gradient problem by decreasing the path length between long range dependencies. Furthermore, compared to RNNs, the computational complexity per layer is less and overall more parallelizable. Specifically, a self attention layer is faster than a recurrent network layer when the sequence length $n$ is less than than the representation's dimensionality $r$, which is often the case. Table 1, modified from \cite{vaswani2017attention} furthers shows this comparison.

\subsection{Explanation of Transformers}
\begin{wrapfigure}{l}{0.4\columnwidth}
  \includegraphics[width=0.4\columnwidth]{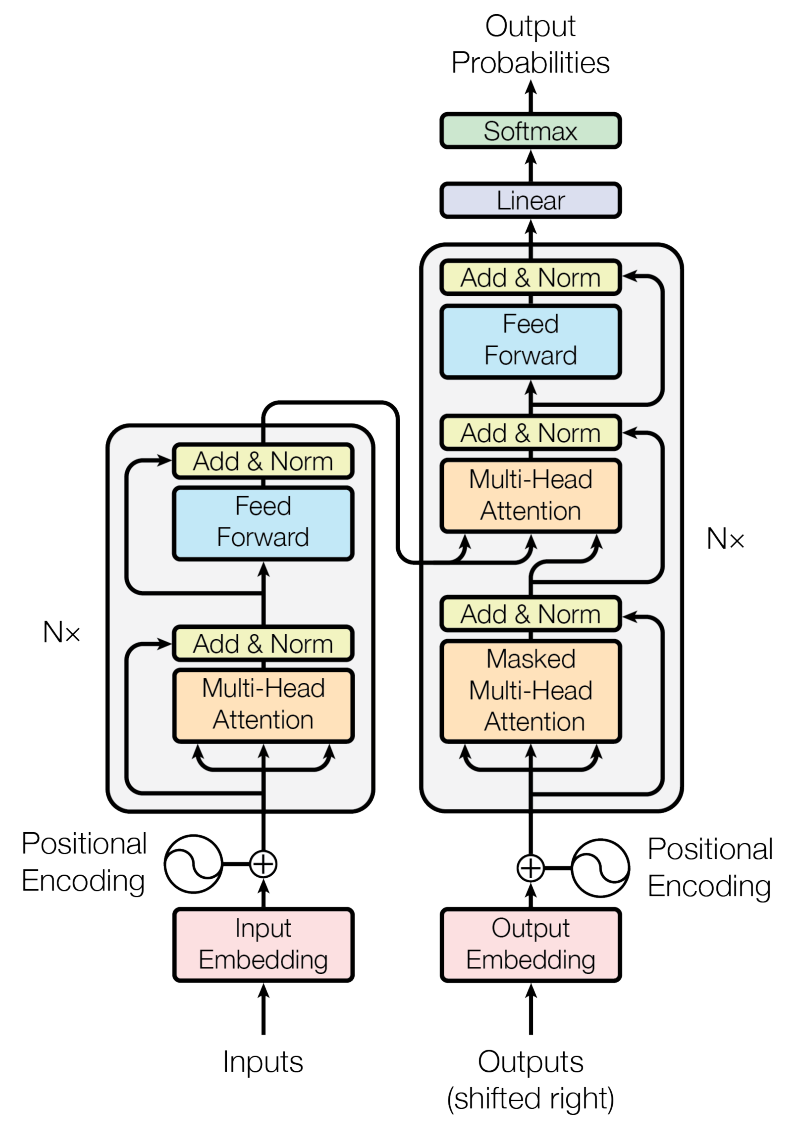}
  \caption{{\bf Transformer Architecture}  Image taken from original CLIP paper, please refer to  \cite{vaswani2017attention} for more details.}
  \label{transformer_fig}
\end{wrapfigure}
A transformer functions through an encoder decoder architecture.
The encoder reads an input sequence simultaneously and creates an intermediate representation. The decoder repeatedly uses that intermediate interpretation as well as the previous outputs of the sequence to produce the next output. The encoder decoder transformer architecture is shown in Fig. \ref{transformer_fig}.

Within the encoder there are self attention modules which compute weighted dot products of each input with each other one. Each input is projected into keys, values, and queries and the values of each input is re-weighted by the soft-max scores of the corresponding query combined with each of the keys. Keys and queries are combined through dot products, but can also be combine through other similarity metrics \cite{vaswani2017attention}. Intuitively, the queries and keys give a higher score where the value needs to pay more attention to. In terms of machine translation, when understanding the meaning of sentence every word does not always provide information to every other word, but some words at arbitrary distances within the sentence will provide more useful meaning. 

Self attention layers treat inputs as an ordered set, i.e. they are permutation equivariant. This helps model long range dependencies in a parallelizable fashion since repeated dot products are repeated matrix multiplications.  However, in many contexts the position of the input in a sequence does matter, so a positional encoding is appended or somehow included in each input vector. Positional encoding can be a fixed function of the input's index in a sequence or a learned function \cite{gehring2017convolutional}.

\subsection{Revolutionary Impacts of Transformers in Language}
Transformers had the largest and most immediate impact in natural language processing (NLP). Before transformers, NLP was dominated by classical methods such as hidden Markov models, conditional random fields, Naive Bayes, and deep learning methods such as RNNs, LSTMs, etc. \cite{khan2016survey}, and now almost every highly used language model is based off of transformers. 

Generative Pre-trained Transformer (GPT) uses the transformer decoder in a autoregressive fashion to develop a robust representation of text that can be used for many NLP tasks\cite{radford2018improving}.
Bidirectional Encoder Representations from Transformers (BERT) similarly learns a representation of natural language, however uses the transformer encoder to encode a sequence of text all at once using context on both sides of the text \cite{devlin2018bert}. Both GPT and BERT follow a pretraining paradigm, where massive models are pretrained in a self-supervised fashion online to create a natural language understanding that can be quickly fine tuned to down stream tasks.

\section{Transformers in Computer Vision}
\label{sec:cv}
Deep learning has arguably dominated computer vision ever since the ImageNet moment in 2015, when intelligent AI models were able to beat standard human recognition ability on image classification. Fueled by increasingly massive datasets, e.g. CIFAR \cite{krizhevsky2009learning}, Imagenet \cite{deng2009imagenet}, COCO \cite{lin2014microsoft},  and massive improvements in GPU computation power, computers are able to more accurately and robustly perform vision tasks. Although transformers were originally designed for language translation and NLP tasks, they have now been adopted for vision tasks and are used as a backbone in many models. 

\begin{figure}
  \includegraphics[width=\columnwidth]{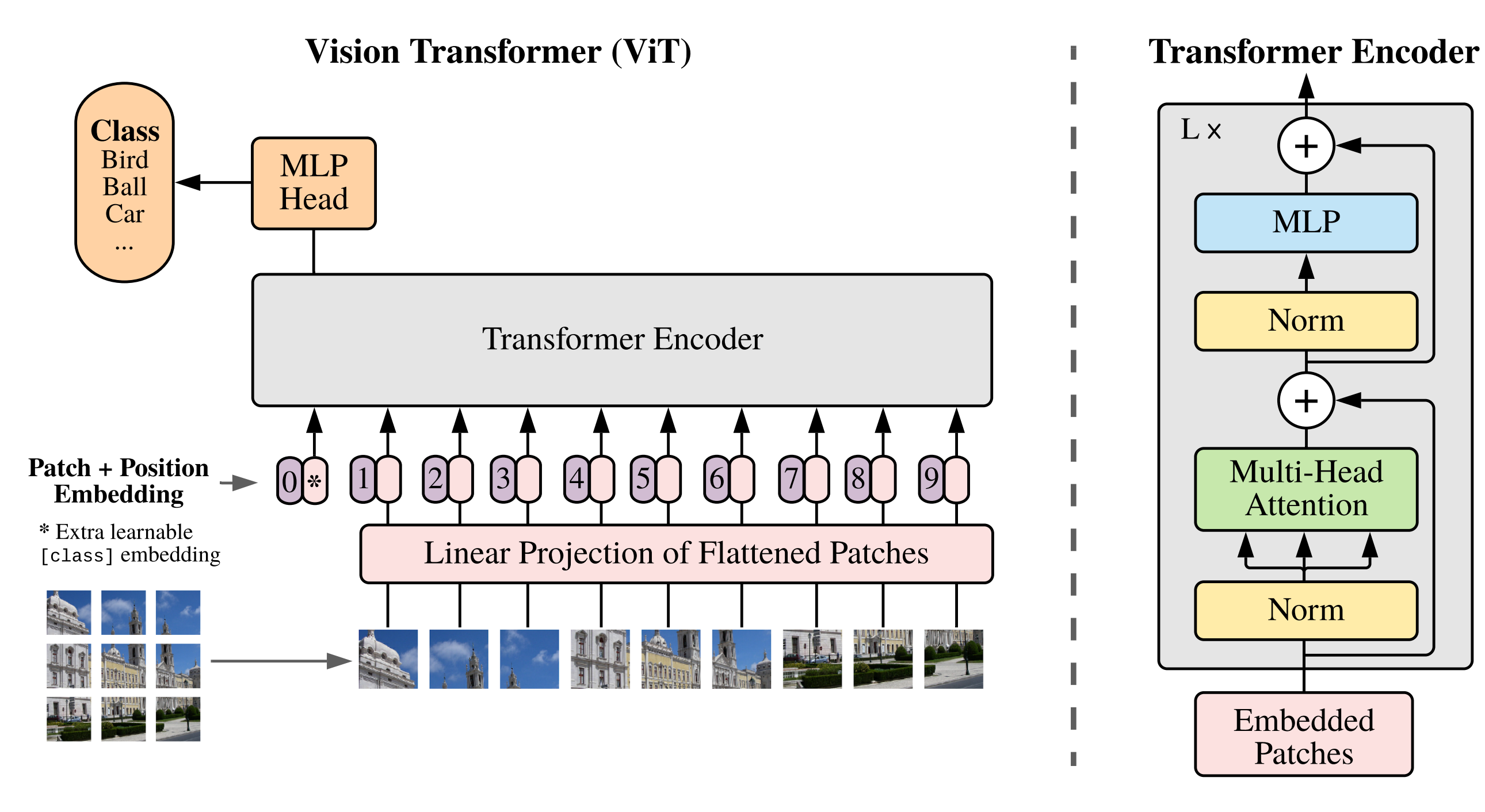}
  \hfill
  \caption{{\bf ViT Overview} ViT splits an image into fixed sizes, embeds each chunk, adds positional encoding and feeds them through a standard transformer. Figure from \cite{dosovitskiy2020image}}
  \label{ViT_fig}
\end{figure}

\subsection{Vision Transformer}
Vision Transformer (ViT) is one of the first and widely used computer vision model that applies transformers successfully \cite{dosovitskiy2020image}. Notably, they are the first work to create a purely transformer based architecture that does not use any convolution layers. Their methodology shown in Fig. \ref{ViT_fig} casts image classification to a sequence-to-sequence problem, by dividing an image into subsquares and treating each as an input token in the sequence. To perform classification they embed an extra token and run its corresponding output through a multi-layer perception (MLP) to determine the class label for the image. 

Dosovitskiy et al. \cite{dosovitskiy2020image} discover that large ViT models performance scales with dataset size more rapidly then resnets. Training a large resenet model on a small dataset performs better than training a large ViT model on a small dataset, however training a large resenet model on a large dataset performs worse than training a large ViT model on a large dataset. This implies that ViT model can capture image representations better across large amounts of data.  In addition, ViT models are more computationally parallelizable and thus require fewer computation resources to train. 

\begin{figure}
  \includegraphics[width=\columnwidth]{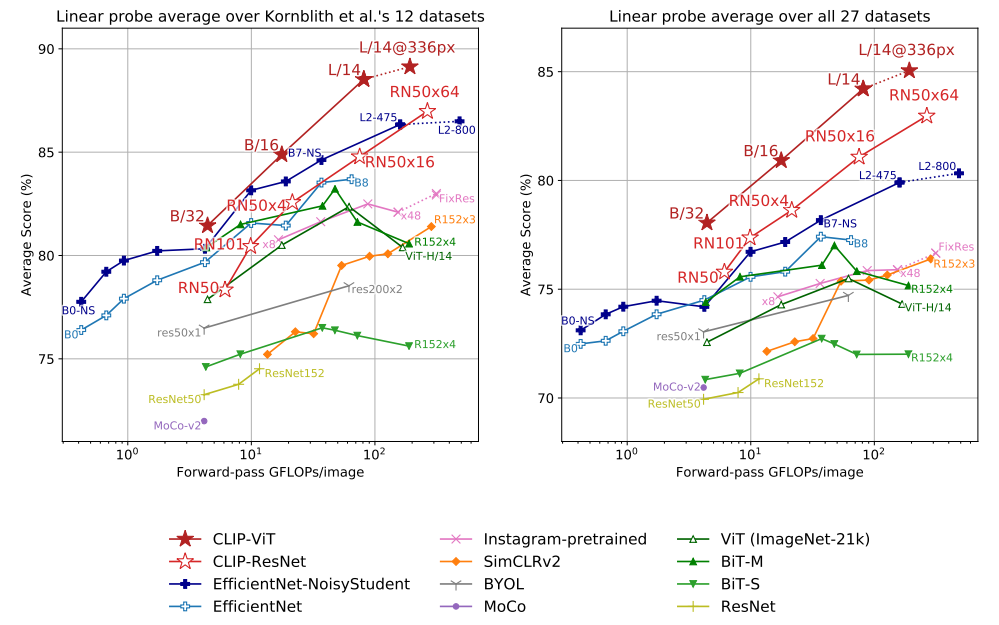}
  \hfill
  \caption{{\bf ViT Performance in CLIP} CLIP trained with a ViT performance best compared to other state of the art models. Image taken from original CLIP paper, please refer to  \cite{radford2021learning} for more details.}
  \label{CLIP_fig}
\end{figure}

ViT's higher representational power and lower computational consumption lends itself to be a strong candidate for large scale visual pretraining tasks. Foundation models is a new term for general multi-modal AI models trained on massive amounts of data in a self-supervised way which aim to obtain a general understanding of the data \cite{bommasani2021opportunities}. Some examples of foundation models include Florence \cite{yuan2021florence}, Dall-E \cite{ramesh2021zero}, CLIP\cite{radford2021learning}, and ALIGN \cite{jia2021scaling}, notably all include a transformer and some even use ViT directly. 

Contrastive Language and Image Pretraing (CLIP) \cite{radford2021learning} uses a technique that utilizes an encoder architecture on both images and text to learn their correlations. They use a massive amount of data (~400 million image-caption pairs) pulled from online and train in a self-supervised fashion called contrastive learning. CLIP beats baseline ResNet50 models in 16 different zero shot learning tasks. As shown in Fig. \ref{CLIP_fig}, CLIP trained with ViT encoders consistently outperforms CLIP trained with ResNet and other convolution based based transfer learning models such as EfficientNet \cite{tan2019efficientnet} or Big Transfer (BiT) \cite{kolesnikov2020big}.

\subsection{DETR}
\begin{figure}
  \includegraphics[width=\columnwidth]{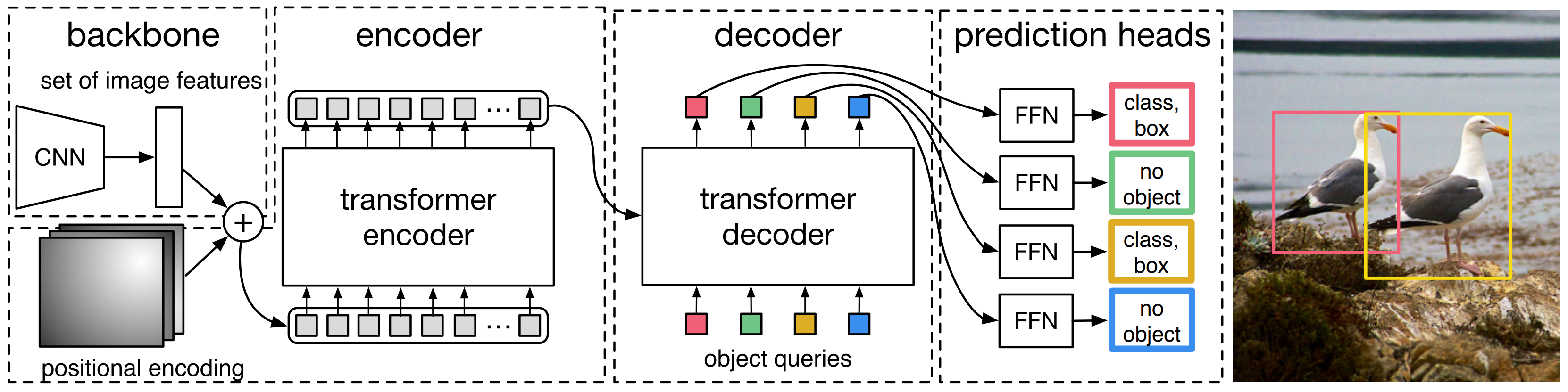}
  \caption{{\bf DETR Architecture} First features are extracted using a CNN backbone, and positional encodings are concatenated to the features. Then features are fed into the encoder and used in the decoded along with learned object queries. Finally, the output s from the decoder are passed through a shared feedforward network that predicts either a detection (class and bounding box) or no detection (through the "no object" class). Image taken from original DETR paper, please refer to \cite{carion2020end} for more details.}
  \label{DETR_fig}
\end{figure}

The aforementioned advances in computer vision include classification and visual image pretraining, however these tasks involve broadly understanding visual representations. Transformers have also proven to effectively perform fine-grained computer vision tasks such as object detection and segmentation. Carion et al. \cite{carion2020end} introduce a new framework called the Detection Transformer (DETR) that views object detection as a set prediction problem. DETR uses a CNN backbone to extract image features, and uses each feature as a token to a standard transformer encoder decoder architecture to produce bounding boxes as illustrated in Fig. \ref{DETR_fig}. One main difference is during the decoder phase, instead of passing decoder outputs back as inputs autoregressively, the decoder learns $n$ object queries to predict $n$ output bounding boxes in parallel through back-propagation through the decoder's cross attention modules. 

\begin{figure}
    \begin{subfigure}{0.44\columnwidth}
  \includegraphics[width=\columnwidth]{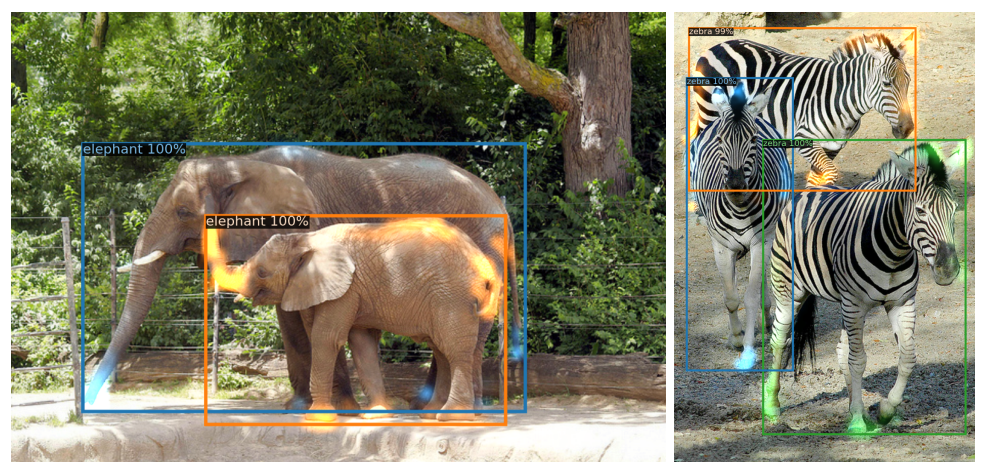}
  \caption{{\bf DETR Attention Visualization} The visualization of the attention scores of the DETR model on the image show that the when determining bounding boxes the model is focusing on the extremities of objects. Image taken from original DETR paper, please refer to \cite{carion2020end} for more details.}
  \label{DETR_attention_fig}
    \end{subfigure}
  \begin{subfigure}{0.5 \columnwidth}
  \includegraphics[width=\columnwidth]{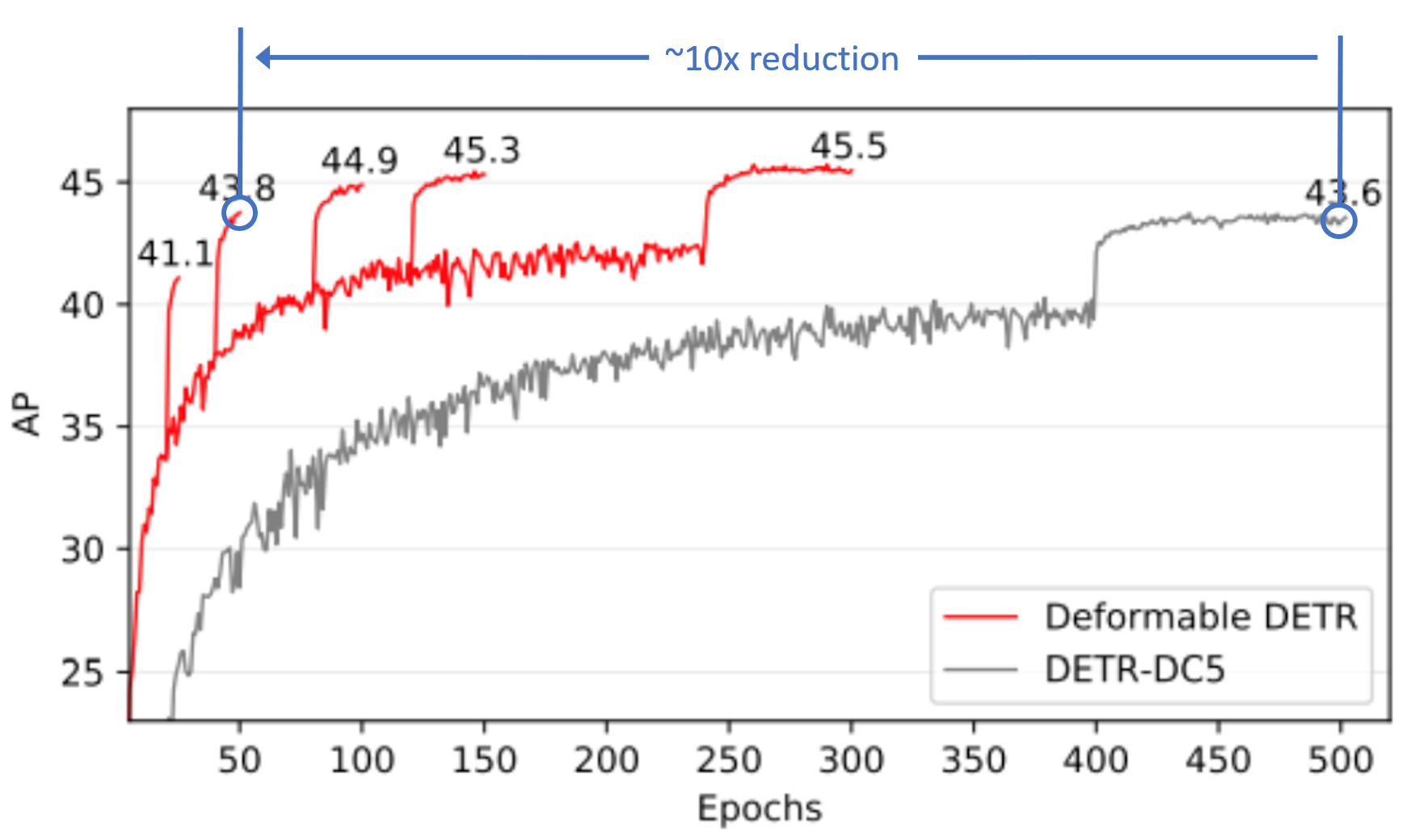}
  \caption{{\bf Deformable DETR}  The graph shows the 10x training speedup of Deformable DETR. Image modified from original Deformable DETR paper, please refer to  \cite{zhu2020deformable} for more details.}
  \label{defDETR_fig}
\end{subfigure}
\vspace{+10pt}
\caption{The left shows visualizations of detection using attention presented in DETR~\cite{carion2020end}. The right shows the speedup of deformable DETR from~\cite{zhu2020deformable}.}
\end{figure}

Surprisingly, DETR did not become well known for its pure performance. In fact, it performs worse on detecting small objects and takes a very long time to train. Nonetheless, it is a simple architecture and it is extensible and relatively interpretable. Compared to other legacy SOTA object detection frameworks, such as Faster R-CNN, DETR does not need a region proposal network (RPN), non maximum suppression and region of interest alignment or any other hand designed components in a detection system. Some works such as YOLO \cite{redmon2016you} and SSD \cite{liu2016ssd} claim to save hand designed additional components (e.g. RPN) for object detection through end-to-end neural network systems, however these works still rely on multiple implicit steps such as bounding box regression and IOU maximum suppression which DETR avoids altogether. Furthermore, through visualizing the transformer attention of the network DETR provides a new interpretable perspective compared to these convolution-based architectures (Fig. \ref{DETR_attention_fig}).

Many works have built upon DETR, one of the most important such extensions is Deformable DETR \cite{zhu2020deformable}. Deformable DETR has shown a 10x significant speedup in training time for similar performance metrics and is thus more widely used in practice . Intuitively, Deformable DETR uses deformable attention layers to learn which neighbors to attend to, thus requiring less inputs to be passed into the transformer over time. The significant speedup is shown in Fig. \ref{defDETR_fig}.

In this section we covered a few of the many impressive works using transformers to outperform existing computer vision tasks. Transformers in computer vision is still a very active research field and is still being adapted in novel ways to different tasks. In the next section we describe the impact transformers are having for one of these tasks, called Multi Object Tracking.

\section{Multi Object Tracking}
\label{sec:mot}
\subsection{Problem description}
Visual Object Tracking (VOT) and Multi Object Tracking (MOT) are an extension of object detection to videos. Visual object tracking is a simpler problem where a detection in the first frame is given as input and the network has to provide a bounding box for that object in subsequent frames. Multi object tracking is a more challenging, yet realistic, extension where the goal is to provide a bounding box, classification label and instance track id for each object in each frame of a video\cite{wang2022recent}. However, it is not as simple as performing frame-by-frame detections since providing a track ID involves determining which objects are the same across frames. This introduces numerous challenges, and 3 of the main challenges are outlined below \cite{xu2019deep}:
\begin{enumerate}
    \item Track creation: During a video a new object can appear in the frames that has not been seen in previous frames. A model must be able to detect it as a new object and create a new track ID, instead of mislabeling it as a previously identified object. This is sometimes referred to as track birth.
    \item Track termination: Objects may leave the frame in the middle of a video and the model must be able to recognize this and terminate that track ID. This is sometimes referred to as track death.
    \item Reidentification (ReID): Objects in videos often get occluded. For example a truck could come in front of a human or a ball could leave the frame bounce and come back. Being able to re-identify an object involves being able to terminate its track ID and revive the same track id when the object returns in the frame instead of creating a new instance label for it.
\end{enumerate}
Other challenges include tracking through occlusions, background clutter and pose changes. Thus, most frameworks perform MOT in 2 high level steps. First, they perform object detection on images and then they associate each of those detections with track IDs across consecutive frames. The detections are canonically performed with pretrained SOTA object detectors, and the associations often use some sort of motion or appearance model.

There are many different applications of MOT such as tracking pedestrians or vehicles for autonomous driving, players on a court for sports analysis, groups of animals in the wilderness for biodiversity studies, etc \cite{luo2021multiple}. As a result, there are many different datasets corresponding to each unique application. 

The most widely used datasets are the MOT challenge datasets, namely MOT16 \cite{milan2016mot16} and it's following iterations (MOT17 and MOT20 \cite{dendorfer2020mot20}). These datasets are mostly surveillance camera footage meant for tracking humans. The authors of the data sets do label cars, bicycles and motorcycles, but only use them as occlusions when calculating the precision and recall of human tracks. 

Another major category of datasets include object tracking for autonomous driving such as Berkely Deep Drive (BDD) \cite{yu2020bdd100k}, and Karlsruhe Institute of Technology and Toyota Technological Institute (KITTI) \cite{Geiger2012CVPR}. Recent works also introduce multi-purpose dataset sets, specifically Track Any Object (TAO) \cite{dave2020tao} has a long-tailed distribution of object classes where some objects appear less frequently than others, and TAO-Open World \cite{liu2022opening} contains novel objects and claims this is more real-world as any model will inevitably run into objects it has never seen before. 

\begin{figure}
  \includegraphics[width=\columnwidth]{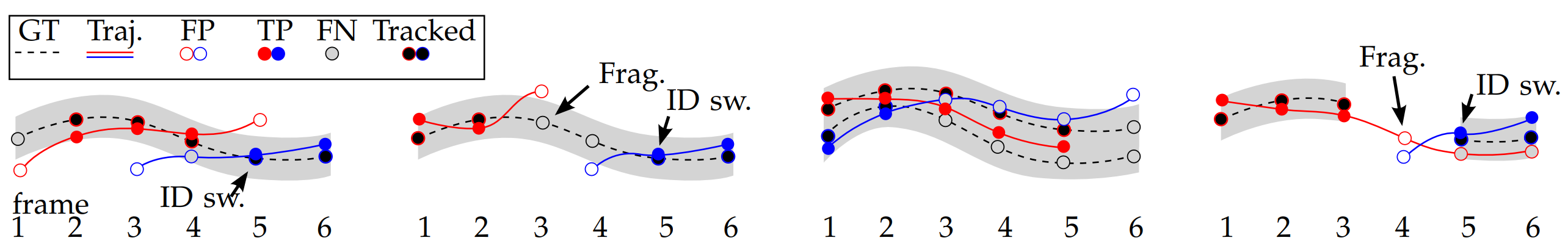}
  \caption{{\bf Fragmentation and IDS in MOT} The figure shows various cases for fragmentions and ID switches. Image taken from original MOT16, please refer to \cite{milan2016mot16} for more details.}
  \label{IDS_fig}
\end{figure}

There are numerous MOT metrics because measuring tracking is a difficult objective. Perhaps the most widely used metric is the multi-object tracking accuracy (MOTA) which corresponds to how well the object can be tracked overall \cite{milan2016mot16}. It combines three error sources: false positives, missed targets and identity switches (IDS). IDS is the number of times an object with a correct track ID switches to an incorrect track ID in the middle of the video. This is correlated with a with another metric called fragmentations (frag) where an object with an existing track ID appears in the image but goes unlabeled by the model or an object with an existing track ID is labeled in an image but the object left the image or is completely occluded (Fig. \ref{IDS_fig}). An IDF1 score is the ratio of correctly identified detections over the average number of ground truth and computed detections. Higher order tracking metric (HOTA) is another recent yet widely used metric that attempts to balance the different aspects of MOT better than MOTA, such as accurate detection, association and localization \cite{luiten2021hota}. 

\subsection{Tracking without Transformers}

Before transformers one of the main MOT methods was Simple Online and Realtime Tracking (SORT) \cite{bewley2016simple}. This was the first method to perform tracking fast enough to be considered realtime. As opposed to previous works that attempted to be robust for all edge cases SORT relies on simple and reliable frame to frame associations. They purposefully ignore ReID since trying to account for it includes unnecessary complexity and precludes real time tracking. Instead they use classical tracking algorithms, such as Kalman filters for state estimate of each of the object IDs and the Hungarian algorithm for data association, with learned object detection models. An extension of SORT, DeepSORT \cite{belmouhcine2021robust} uses object appearances as well to improve ReID of a person on the MOT17 dataset. They also train a CNN to discriminate pedestrians offline, and apply it during the association step to decrease the IDS metric.

\begin{figure}
  \includegraphics[width=\columnwidth]{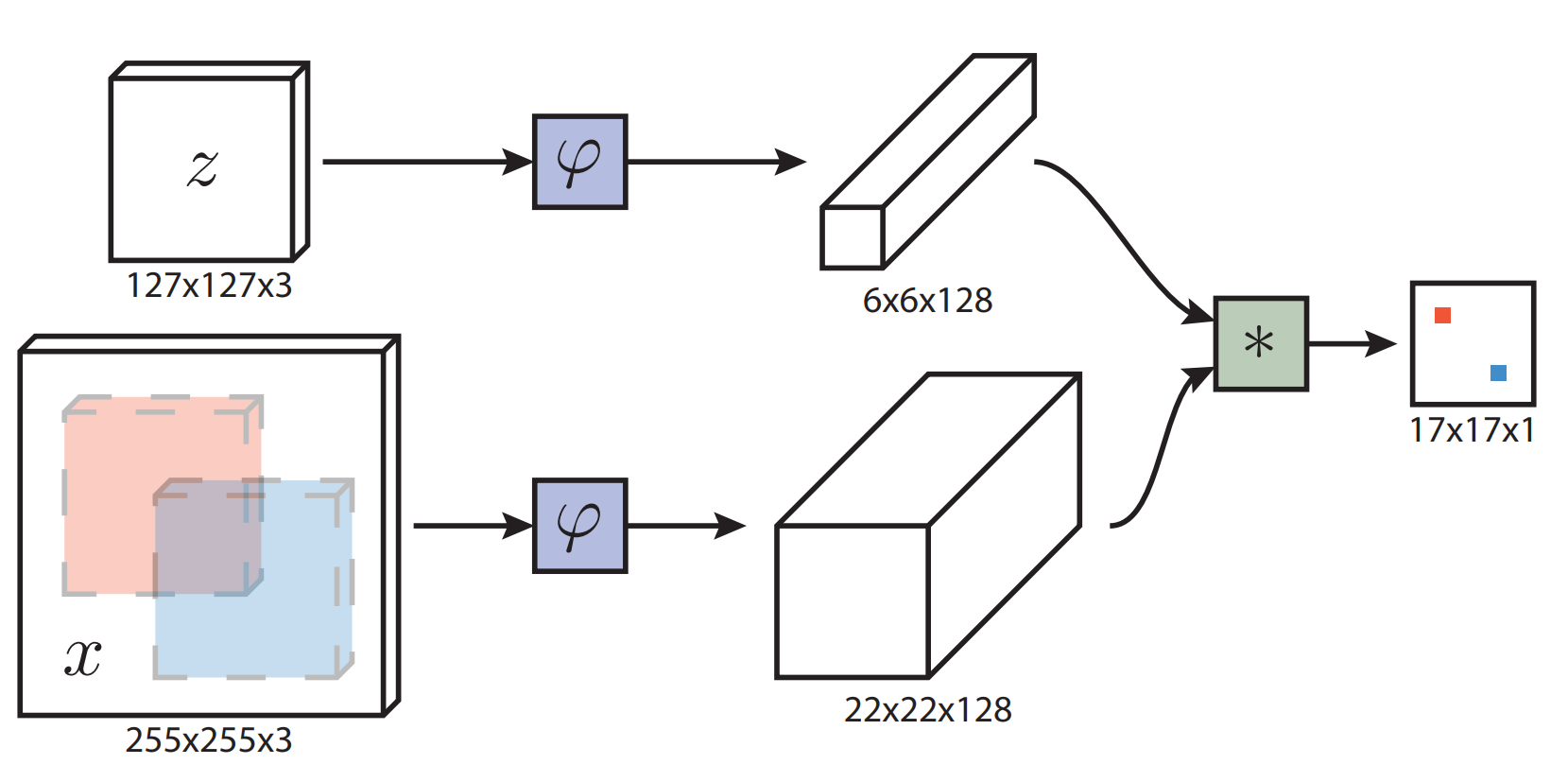}
  \caption{{\bf Siamese Network Concept} The figure shows simple explanation of the concept behind a Siamese Network. The detection $z$ from image $x$ is passed through a different branch then the rest of the image and the branches output a filter which is convolved as a template with the original images feature map. Now $x$ can be updated to be the next image in the frame while $z$ remains to determine where that object track ID appears in the next image. Image taken from original SiamFC, please refer to \cite{bertinetto2016fully} for more details.}
  \label{siam_fig}
\end{figure}

Around the same time as SORT, another common tracking method was through siamese networks such as SiamFC \cite{bertinetto2016fully}. Siamese networks include 2 branches, one that learns the canonical object detection used for tracking and another that learns to associate objects through frames. The goal is to learn convolutional features of an object in one detection that can be used as a template or convolutional filter itself in the other branch. SiamFC refers to this second branch as similarity learning, where they try to learn a network that can tell if two objects are similar and gives it a higher score if so. The detections in the first branch essentially serve as "exemplar images" for the second branch to predict where that exact object was, by applying the learned filter or template for that object to all parts of the image (like a regular convolution) (Fig. \ref{siam_fig}). 

SiamRPN \cite{li2018high} uses a similar idea of having one detection as a template for other objects on that track by computing the correlation between the template and the detections in the following images. This formulates the tracking problem as a one shot detection problem: use the template branch to predict weights for kernels in the RPN of an object detector and regress accordingly on those results. During inference, only the first frame runs through both branches then the template branch is turned off and the weights outputted from it are used in the RPN for the rest of the images detectors to perform tracking. 

A work called Joint Detection and Embedding (JDE) \cite{wang2020towards} noticed that previous tracking works have two stages, one was outputting bounding boxes and features and the second is passing those detections through a separate ReID network to correlate "appearance features" to label track IDs. JDE creates an end to end network to determine both object detections and object appearance embeddings at once. They test on MOT16 and don't out perform SOTA methods on the MOTA metric at their time, however, they get really close and perform 3x faster, at 24 FPS "near real-time."

Traktor \cite{bergmann2019tracking} takes advantage of object detection RPN and bounding box regression methods to do tracking. They predict the object ID in the next frame by regressing previous bounding boxes on the current frame's features maps to predict track IDs. They further provide an extension of their mode, Traktor++ which adds a ReID siamese network and motion model to perform association better. Traktor++ work achieved SOTA results on MOT15,16, and 17 at that time (2019).

CenterTrack \cite{zhou2020tracking} is based off of a novel detection framework that detects objects as points (CenterNet\cite{zhou2019objects}). Tracking objects as points simplifies tracking conditioned detection as many objects can be represented as heatmaps of points. This condensed representation allows the next frame's detector to be condition on multiple previous frames, which works well especially for low frame rate sequences where boxes between frames might not overlap, compared to detectors like Traktor \cite{bergmann2019tracking} which relies on the overlap in the RPN to link detections into tracks. This is one of the first tracking works that trains on static images with aggressive data augmentation, i.e. they crop and shift detections to mimic motion and expand their training dataset.

\cite{tokmakov2021learning} extends on Centertrack\cite{zhou2020tracking}, specifically focusing on tracking through occlusions. Working off of KITTI, they claim to track pedestrians that are fully occluded by a vehicle. They do such by augmenting a recurrent memory module RNN to reason about object location and identity. Their custom synthetic ParallelDomain dataset they train on has groundtruth labels behind complete occlusions which warrants their success on KITTI.

AOA \cite{du20211st} is the first place winner in the TAO challenge. They argue that MOT trackers often learn both appearance and movement models but object movement is too unpredictable. They create a model that tracks objects based only on appearance. This aligns with their success in the Track Any Object Dataset because a ball or bird moves very differently from a car or person and predicting a general motion model for any object may lead to more errors than benefits. They use an ensemble of detection and ReID networks to output multiple tracks which are then averaged or merged together.

Recently, a few more improved SORT methods have arisen. StrongSORT \cite{du2022strongsort} improves DeepSORT with an appearance-free link to use only spatio-temporal information to predict whether tracklets (snippets of tracks across multiple frames) belong to the same ID or not. They also use guassian smoothed interpolation as opposed to linear interpolations for missing detections. OC-SORT \cite{cao2022observation}, claims most models assume simple and linear motion, but instead if you have a more robust motion model you can perform well without an appearance model through occlusions and nonlinear motions. This especially performed well on the DanceTrack dataset where performers had similar costumes and appearances.

ByteTrack \cite{zhang2022bytetrack} is a more recent SOTA method that attempts to revive dropped or fragmented tracks. Their intuition follows that when an object is occluded or blurred through motion its bounding box prediction is going to be less confident, yet the detection should not be discarded. Instead ByteTrack performs two rounds of associations, where it firsts associates track IDs to objects with high confidence detection results, and then associates the remaining track IDs to low confidence detections. As per its original motivation, ByteTrack achieved SOTA IDF1 and IDS scores when it came out.

\subsection{Tracking with Transformers}
\begin{wrapfigure}{L}{0.4\columnwidth}
  \includegraphics[width=0.4\columnwidth]{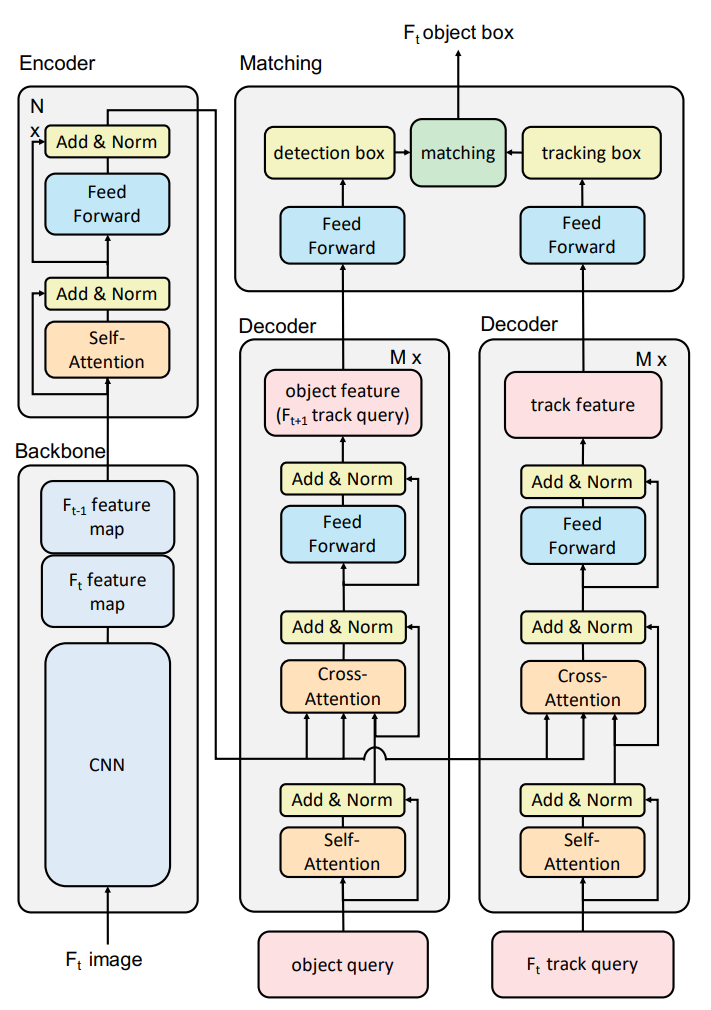}
  \caption{{\bf TransTrack Architecture}  Image taken from original TransTrack paper, please refer to  \cite{sun2020transtrack} for more details.}
  \label{transtrack_fig}
  \vspace{-.5cm}
\end{wrapfigure}

Soft data association (SoDA) \cite{hung2020soda}  uses attention to encode spatio-temporal dependencies and better track through occlusions, as shown by their low IDS score on the MOT17 dataset.  To our knowledge, TransTrack (2020)~\cite{sun2020transtrack} is the first work to successfully apply the transformer model to tracking. They apply object features from a previous frame as a query for the current frame and introduce the concept of track queries that are inputs to a decoder representing existing tracks. They still use the concept of object queries from DETR in a second decoder. The network essentially associates the track queries with detected objects from the object queries. The complete architecture is shown in Fig. \ref{transtrack_fig}. Similar to TransTrack, Trackformer\cite{meinhardt2022trackformer} extends DETR almost directly using object and Track queries. Trackformer only has one decoder which does the heavy lifting part of the tracking, including the birth and death of tracks. Multiple-Object Tracking with Transformers (MOTR) \cite{zeng2022motr}, extends Trackformer with a Query Interaction Module which interacts with the decoder and more precisely handles the birth and death of track queries.

TransMOT \cite{chu2021transmot} points to issues in the TransTrack \cite{sun2020transtrack} line of work being that these works do not account for the spatial temporal structure of the object, and they require lots of computation. TransMOT represents detections in a frame as a graph and then use a transformer made from graph neural networks to create tracks. 

Transcenter \cite{xu2021transcenter} argues existing tracking with transformers works are un-optimal given that their initial learned object and track queries are initialized by noise. They train a query learning network to get image-related dense prediction queries and sparse tracking queries and use those to perform tracking better.

\begin{figure}
  \includegraphics[width=\columnwidth]{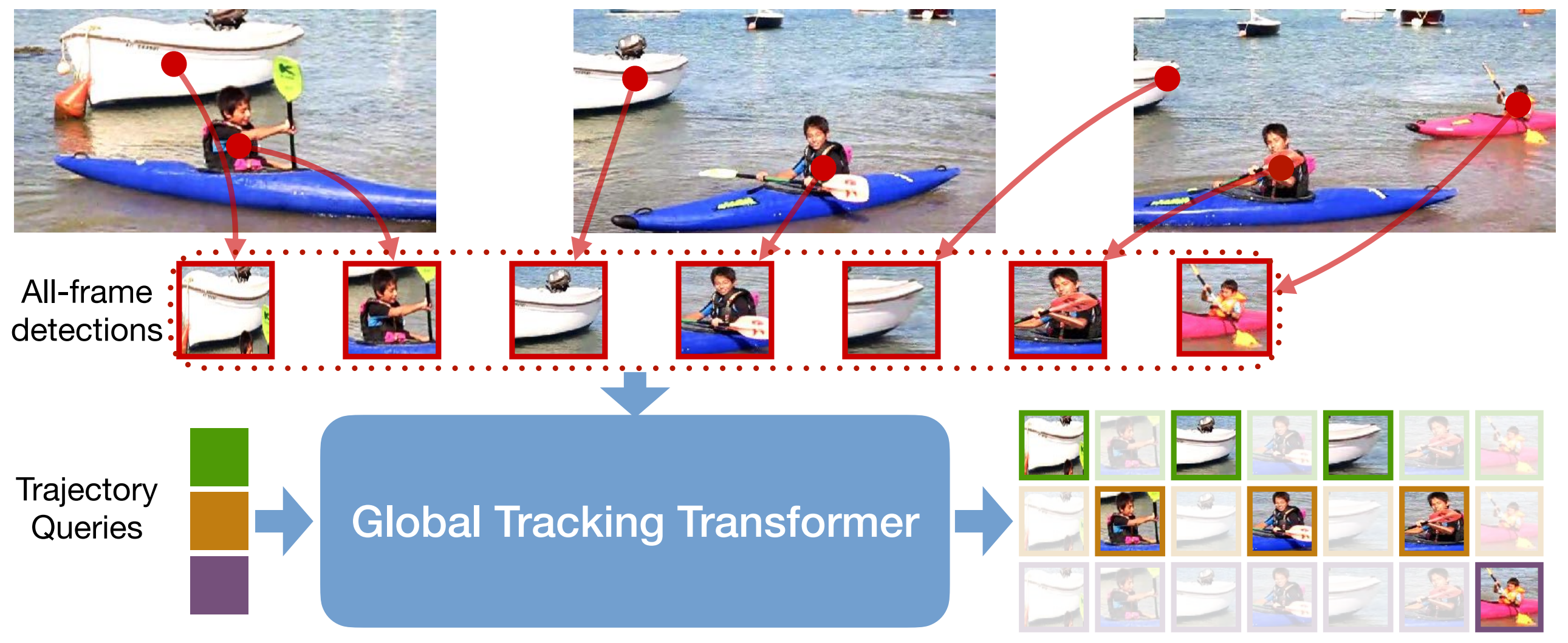}
  \caption{{\bf Global Tracking Transformer} Image taken from original GTR paper, please refer to \cite{zhou2022global} for more details.}
  \label{gtr_fig}
\end{figure}

Global Tracking Transformers (GTR) \cite{zhou2022global} provides a novel approach by using transformers only for tracking, instead of joint detection and tracking. They perform object detection first and send each of the detection crops across a temporal window of frames into a transformer. Using learned trajectory-queries, similar to track-queries in previous works, the transformer outputs a vector for each query scoring how well each input detection belongs to the corresponding query's trackID (Fig. \ref{gtr_fig}). This architecture is simple and lightweight and gains a lot of information from using a temporal window as opposed to simply the past 1 or 2 frames, however, it does not perform state of the art.

\subsection{State of the Art Tracking}
Although transformers have performed extremely well on many tasks, whether they will outperform existing methods in object tracking is still uncertain. Papers with Code is a website that ranks existing works faster than they can be published for ongoing research tasks. Observing the MOT17 benchmark on Papers with Code \footnote{https://paperswithcode.com/sota/multi-object-tracking-on-mot17}, the most widely used MOT benchmark, we see the top two performers are BoT-SORT \cite{aharon2022bot}, and SMILEtrack \cite{wang2022smiletrack}. BoT-SORT combines motion, appearance, camera-motion and Kalman filters to improve upon ByteTrack and the original SORT algorithm. SMLE uses a similarity learning module motivated from Siamese networks to correlate objects to tracks. Ironically, the top SOTA models today are not one of the many transformer-related tracking papers but instead are using the first 2 foundational MOT frameworks we discussed above from about 7 years ago, SORT\cite{bewley2016simple} and SiamFC\cite{bertinetto2016fully}.

\section{Conclusion}
We have presented an extensive review of transformers in computer vision. We discuss the background of transformers and how they have impacted NLP. We further describe how they made their way into computer vision and are having a tranformative impact on SOTA pretraining models and object detection. Finally, we dive into the field of Multi-Object Tracking and show how transformers are playing a role there. Although, transformers have performed well in MOT occasionally, current SOTA does not use transformers and the field is highly active.

{\small
\bibliographystyle{ieee_fullname}
\bibliography{refs}

\begin{thebibliography}{10}\itemsep=-1pt

\bibitem{aharon2022bot}
Nir Aharon, Roy Orfaig, and Ben-Zion Bobrovsky.
\newblock Bot-sort: Robust associations multi-pedestrian tracking.
\newblock {\em arXiv preprint arXiv:2206.14651}, 2022.

\bibitem{belmouhcine2021robust}
Abdelbadie Belmouhcine, Julien Simon, Luc Courtrai, and S{\'e}bastien
  Lef{\`e}vre.
\newblock Robust deep simple online real-time tracking.
\newblock In {\em 2021 12th International Symposium on Image and Signal
  Processing and Analysis (ISPA)}, pages 138--144. IEEE, 2021.

\bibitem{bergmann2019tracking}
Philipp Bergmann, Tim Meinhardt, and Laura Leal-Taixe.
\newblock Tracking without bells and whistles.
\newblock In {\em Proceedings of the IEEE/CVF International Conference on
  Computer Vision}, pages 941--951, 2019.

\bibitem{bertinetto2016fully}
Luca Bertinetto, Jack Valmadre, Joao~F Henriques, Andrea Vedaldi, and Philip~HS
  Torr.
\newblock Fully-convolutional siamese networks for object tracking.
\newblock In {\em European conference on computer vision}, pages 850--865.
  Springer, 2016.

\bibitem{bewley2016simple}
Alex Bewley, Zongyuan Ge, Lionel Ott, Fabio Ramos, and Ben Upcroft.
\newblock Simple online and realtime tracking.
\newblock In {\em 2016 IEEE international conference on image processing
  (ICIP)}, pages 3464--3468. IEEE, 2016.

\bibitem{bommasani2021opportunities}
Rishi Bommasani, Drew~A Hudson, Ehsan Adeli, Russ Altman, Simran Arora, Sydney
  von Arx, Michael~S Bernstein, Jeannette Bohg, Antoine Bosselut, Emma
  Brunskill, et~al.
\newblock On the opportunities and risks of foundation models.
\newblock {\em arXiv preprint arXiv:2108.07258}, 2021.

\bibitem{cao2022observation}
Jinkun Cao, Xinshuo Weng, Rawal Khirodkar, Jiangmiao Pang, and Kris Kitani.
\newblock Observation-centric sort: Rethinking sort for robust multi-object
  tracking.
\newblock {\em arXiv preprint arXiv:2203.14360}, 2022.

\bibitem{carion2020end}
Nicolas Carion, Francisco Massa, Gabriel Synnaeve, Nicolas Usunier, Alexander
  Kirillov, and Sergey Zagoruyko.
\newblock End-to-end object detection with transformers.
\newblock In {\em European conference on computer vision}, pages 213--229.
  Springer, 2020.

\bibitem{chu2021transmot}
Peng Chu, Jiang Wang, Quanzeng You, Haibin Ling, and Zicheng Liu.
\newblock Transmot: Spatial-temporal graph transformer for multiple object
  tracking.
\newblock {\em arXiv preprint arXiv:2104.00194}, 2021.

\bibitem{chung2014empirical}
Junyoung Chung, Caglar Gulcehre, KyungHyun Cho, and Yoshua Bengio.
\newblock Empirical evaluation of gated recurrent neural networks on sequence
  modeling.
\newblock {\em arXiv preprint arXiv:1412.3555}, 2014.

\bibitem{dave2020tao}
Achal Dave, Tarasha Khurana, Pavel Tokmakov, Cordelia Schmid, and Deva Ramanan.
\newblock Tao: A large-scale benchmark for tracking any object.
\newblock In {\em European conference on computer vision}, pages 436--454.
  Springer, 2020.

\bibitem{dendorfer2020mot20}
Patrick Dendorfer, Hamid Rezatofighi, Anton Milan, Javen Shi, Daniel Cremers,
  Ian Reid, Stefan Roth, Konrad Schindler, and Laura Leal-Taix{\'e}.
\newblock Mot20: A benchmark for multi object tracking in crowded scenes.
\newblock {\em arXiv preprint arXiv:2003.09003}, 2020.

\bibitem{deng2009imagenet}
Jia Deng, Wei Dong, Richard Socher, Li-Jia Li, Kai Li, and Li Fei-Fei.
\newblock Imagenet: A large-scale hierarchical image database.
\newblock In {\em 2009 IEEE conference on computer vision and pattern
  recognition}, pages 248--255. Ieee, 2009.

\bibitem{devlin2018bert}
Jacob Devlin, Ming-Wei Chang, Kenton Lee, and Kristina Toutanova.
\newblock Bert: Pre-training of deep bidirectional transformers for language
  understanding.
\newblock {\em arXiv preprint arXiv:1810.04805}, 2018.

\bibitem{dosovitskiy2020image}
Alexey Dosovitskiy, Lucas Beyer, Alexander Kolesnikov, Dirk Weissenborn,
  Xiaohua Zhai, Thomas Unterthiner, Mostafa Dehghani, Matthias Minderer, Georg
  Heigold, Sylvain Gelly, et~al.
\newblock An image is worth 16x16 words: Transformers for image recognition at
  scale.
\newblock {\em arXiv preprint arXiv:2010.11929}, 2020.

\bibitem{du20211st}
Fei Du, Bo Xu, Jiasheng Tang, Yuqi Zhang, Fan Wang, and Hao Li.
\newblock 1st place solution to eccv-tao-2020: Detect and represent any object
  for tracking.
\newblock {\em arXiv preprint arXiv:2101.08040}, 2021.

\bibitem{du2022strongsort}
Yunhao Du, Yang Song, Bo Yang, and Yanyun Zhao.
\newblock Strongsort: Make deepsort great again.
\newblock {\em arXiv preprint arXiv:2202.13514}, 2022.

\bibitem{elman1990finding}
Jeffrey~L Elman.
\newblock Finding structure in time.
\newblock {\em Cognitive science}, 14(2):179--211, 1990.

\bibitem{gehring2017convolutional}
Jonas Gehring, Michael Auli, David Grangier, Denis Yarats, and Yann~N Dauphin.
\newblock Convolutional sequence to sequence learning.
\newblock In {\em International conference on machine learning}, pages
  1243--1252. PMLR, 2017.

\bibitem{Geiger2012CVPR}
Andreas Geiger, Philip Lenz, and Raquel Urtasun.
\newblock Are we ready for autonomous driving? the kitti vision benchmark
  suite.
\newblock In {\em Conference on Computer Vision and Pattern Recognition
  (CVPR)}, 2012.

\bibitem{hochreiter1998vanishing}
Sepp Hochreiter.
\newblock The vanishing gradient problem during learning recurrent neural nets
  and problem solutions.
\newblock {\em International Journal of Uncertainty, Fuzziness and
  Knowledge-Based Systems}, 6(02):107--116, 1998.

\bibitem{hochreiter1997long}
Sepp Hochreiter and J{\"u}rgen Schmidhuber.
\newblock Long short-term memory.
\newblock {\em Neural computation}, 9(8):1735--1780, 1997.

\bibitem{hung2020soda}
Wei-Chih Hung, Henrik Kretzschmar, Tsung-Yi Lin, Yuning Chai, Ruichi Yu,
  Ming-Hsuan Yang, and Dragomir Anguelov.
\newblock Soda: Multi-object tracking with soft data association.
\newblock {\em arXiv preprint arXiv:2008.07725}, 2020.

\bibitem{jia2021scaling}
Chao Jia, Yinfei Yang, Ye Xia, Yi-Ting Chen, Zarana Parekh, Hieu Pham, Quoc Le,
  Yun-Hsuan Sung, Zhen Li, and Tom Duerig.
\newblock Scaling up visual and vision-language representation learning with
  noisy text supervision.
\newblock In {\em International Conference on Machine Learning}, pages
  4904--4916. PMLR, 2021.

\bibitem{khan2016survey}
Wahab Khan, Ali Daud, Jamal~A Nasir, and Tehmina Amjad.
\newblock A survey on the state-of-the-art machine learning models in the
  context of nlp.
\newblock {\em Kuwait journal of Science}, 43(4), 2016.

\bibitem{kolesnikov2020big}
Alexander Kolesnikov, Lucas Beyer, Xiaohua Zhai, Joan Puigcerver, Jessica Yung,
  Sylvain Gelly, and Neil Houlsby.
\newblock Big transfer (bit): General visual representation learning.
\newblock In {\em European conference on computer vision}, pages 491--507.
  Springer, 2020.

\bibitem{krizhevsky2009learning}
Alex Krizhevsky, Geoffrey Hinton, et~al.
\newblock Learning multiple layers of features from tiny images.
\newblock 2009.

\bibitem{li2018high}
Bo Li, Junjie Yan, Wei Wu, Zheng Zhu, and Xiaolin Hu.
\newblock High performance visual tracking with siamese region proposal
  network.
\newblock In {\em Proceedings of the IEEE conference on computer vision and
  pattern recognition}, pages 8971--8980, 2018.

\bibitem{lin2014microsoft}
Tsung-Yi Lin, Michael Maire, Serge Belongie, James Hays, Pietro Perona, Deva
  Ramanan, Piotr Doll{\'a}r, and C~Lawrence Zitnick.
\newblock Microsoft coco: Common objects in context.
\newblock In {\em European conference on computer vision}, pages 740--755.
  Springer, 2014.

\bibitem{liu2016ssd}
Wei Liu, Dragomir Anguelov, Dumitru Erhan, Christian Szegedy, Scott Reed,
  Cheng-Yang Fu, and Alexander~C Berg.
\newblock Ssd: Single shot multibox detector.
\newblock In {\em European conference on computer vision}, pages 21--37.
  Springer, 2016.

\bibitem{liu2022opening}
Yang Liu, Idil~Esen Zulfikar, Jonathon Luiten, Achal Dave, Deva Ramanan,
  Bastian Leibe, Aljo{\v{s}}a O{\v{s}}ep, and Laura Leal-Taix{\'e}.
\newblock Opening up open world tracking.
\newblock In {\em Proceedings of the IEEE/CVF Conference on Computer Vision and
  Pattern Recognition}, pages 19045--19055, 2022.

\bibitem{luiten2021hota}
Jonathon Luiten, Aljosa Osep, Patrick Dendorfer, Philip Torr, Andreas Geiger,
  Laura Leal-Taix{\'e}, and Bastian Leibe.
\newblock Hota: A higher order metric for evaluating multi-object tracking.
\newblock {\em International journal of computer vision}, 129(2):548--578,
  2021.

\bibitem{luo2021multiple}
Wenhan Luo, Junliang Xing, Anton Milan, Xiaoqin Zhang, Wei Liu, and Tae-Kyun
  Kim.
\newblock Multiple object tracking: A literature review.
\newblock {\em Artificial Intelligence}, 293:103448, 2021.

\bibitem{meinhardt2022trackformer}
Tim Meinhardt, Alexander Kirillov, Laura Leal-Taixe, and Christoph
  Feichtenhofer.
\newblock Trackformer: Multi-object tracking with transformers.
\newblock In {\em Proceedings of the IEEE/CVF Conference on Computer Vision and
  Pattern Recognition}, pages 8844--8854, 2022.

\bibitem{milan2016mot16}
Anton Milan, Laura Leal-Taix{\'e}, Ian Reid, Stefan Roth, and Konrad Schindler.
\newblock Mot16: A benchmark for multi-object tracking.
\newblock {\em arXiv preprint arXiv:1603.00831}, 2016.

\bibitem{radford2021learning}
Alec Radford, Jong~Wook Kim, Chris Hallacy, Aditya Ramesh, Gabriel Goh,
  Sandhini Agarwal, Girish Sastry, Amanda Askell, Pamela Mishkin, Jack Clark,
  et~al.
\newblock Learning transferable visual models from natural language
  supervision.
\newblock In {\em International Conference on Machine Learning}, pages
  8748--8763. PMLR, 2021.

\bibitem{radford2018improving}
Alec Radford, Karthik Narasimhan, Tim Salimans, Ilya Sutskever, et~al.
\newblock Improving language understanding by generative pre-training.
\newblock 2018.

\bibitem{ramesh2021zero}
Aditya Ramesh, Mikhail Pavlov, Gabriel Goh, Scott Gray, Chelsea Voss, Alec
  Radford, Mark Chen, and Ilya Sutskever.
\newblock Zero-shot text-to-image generation.
\newblock In {\em International Conference on Machine Learning}, pages
  8821--8831. PMLR, 2021.

\bibitem{redmon2016you}
Joseph Redmon, Santosh Divvala, Ross Girshick, and Ali Farhadi.
\newblock You only look once: Unified, real-time object detection.
\newblock In {\em Proceedings of the IEEE conference on computer vision and
  pattern recognition}, pages 779--788, 2016.

\bibitem{sun2020transtrack}
Peize Sun, Jinkun Cao, Yi Jiang, Rufeng Zhang, Enze Xie, Zehuan Yuan, Changhu
  Wang, and Ping Luo.
\newblock Transtrack: Multiple object tracking with transformer.
\newblock {\em arXiv preprint arXiv:2012.15460}, 2020.

\bibitem{tan2019efficientnet}
Mingxing Tan and Quoc Le.
\newblock Efficientnet: Rethinking model scaling for convolutional neural
  networks.
\newblock In {\em International conference on machine learning}, pages
  6105--6114. PMLR, 2019.

\bibitem{tokmakov2021learning}
Pavel Tokmakov, Jie Li, Wolfram Burgard, and Adrien Gaidon.
\newblock Learning to track with object permanence.
\newblock In {\em Proceedings of the IEEE/CVF International Conference on
  Computer Vision}, pages 10860--10869, 2021.

\bibitem{vaswani2017attention}
Ashish Vaswani, Noam Shazeer, Niki Parmar, Jakob Uszkoreit, Llion Jones,
  Aidan~N Gomez, {\L}ukasz Kaiser, and Illia Polosukhin.
\newblock Attention is all you need.
\newblock {\em Advances in neural information processing systems}, 30, 2017.

\bibitem{wang2022recent}
Gaoang Wang, Mingli Song, and Jenq-Neng Hwang.
\newblock Recent advances in embedding methods for multi-object tracking: A
  survey.
\newblock {\em arXiv preprint arXiv:2205.10766}, 2022.

\bibitem{wang2022smiletrack}
Yu-Hsiang Wang.
\newblock Smiletrack: Similarity learning for multiple object tracking.
\newblock {\em arXiv preprint arXiv:2211.08824}, 2022.

\bibitem{wang2020towards}
Zhongdao Wang, Liang Zheng, Yixuan Liu, Yali Li, and Shengjin Wang.
\newblock Towards real-time multi-object tracking.
\newblock In {\em European Conference on Computer Vision}, pages 107--122.
  Springer, 2020.

\bibitem{xu2021transcenter}
Yihong Xu, Yutong Ban, Guillaume Delorme, Chuang Gan, Daniela Rus, and Xavier
  Alameda-Pineda.
\newblock Transcenter: Transformers with dense queries for multiple-object
  tracking.
\newblock {\em arXiv preprint arXiv:2103.15145}, 2021.

\bibitem{xu2019deep}
Yingkun Xu, Xiaolong Zhou, Shengyong Chen, and Fenfen Li.
\newblock Deep learning for multiple object tracking: a survey.
\newblock {\em IET Computer Vision}, 13(4):355--368, 2019.

\bibitem{yu2020bdd100k}
Fisher Yu, Haofeng Chen, Xin Wang, Wenqi Xian, Yingying Chen, Fangchen Liu,
  Vashisht Madhavan, and Trevor Darrell.
\newblock Bdd100k: A diverse driving dataset for heterogeneous multitask
  learning.
\newblock In {\em Proceedings of the IEEE/CVF conference on computer vision and
  pattern recognition}, pages 2636--2645, 2020.

\bibitem{yuan2021florence}
Lu Yuan, Dongdong Chen, Yi-Ling Chen, Noel Codella, Xiyang Dai, Jianfeng Gao,
  Houdong Hu, Xuedong Huang, Boxin Li, Chunyuan Li, et~al.
\newblock Florence: A new foundation model for computer vision.
\newblock {\em arXiv preprint arXiv:2111.11432}, 2021.

\bibitem{zeng2022motr}
Fangao Zeng, Bin Dong, Yuang Zhang, Tiancai Wang, Xiangyu Zhang, and Yichen
  Wei.
\newblock Motr: End-to-end multiple-object tracking with transformer.
\newblock In {\em European Conference on Computer Vision}, pages 659--675.
  Springer, 2022.

\bibitem{zhang2022bytetrack}
Yifu Zhang, Peize Sun, Yi Jiang, Dongdong Yu, Fucheng Weng, Zehuan Yuan, Ping
  Luo, Wenyu Liu, and Xinggang Wang.
\newblock Bytetrack: Multi-object tracking by associating every detection box.
\newblock In {\em European Conference on Computer Vision}, pages 1--21.
  Springer, 2022.

\bibitem{zhou2020tracking}
Xingyi Zhou, Vladlen Koltun, and Philipp Kr{\"a}henb{\"u}hl.
\newblock Tracking objects as points.
\newblock In {\em European Conference on Computer Vision}, pages 474--490.
  Springer, 2020.

\bibitem{zhou2019objects}
Xingyi Zhou, Dequan Wang, and Philipp Kr{\"a}henb{\"u}hl.
\newblock Objects as points.
\newblock {\em arXiv preprint arXiv:1904.07850}, 2019.

\bibitem{zhou2022global}
Xingyi Zhou, Tianwei Yin, Vladlen Koltun, and Philipp Kr{\"a}henb{\"u}hl.
\newblock Global tracking transformers.
\newblock In {\em Proceedings of the IEEE/CVF Conference on Computer Vision and
  Pattern Recognition}, pages 8771--8780, 2022.

\bibitem{zhu2020deformable}
Xizhou Zhu, Weijie Su, Lewei Lu, Bin Li, Xiaogang Wang, and Jifeng Dai.
\newblock Deformable detr: Deformable transformers for end-to-end object
  detection.
\newblock {\em arXiv preprint arXiv:2010.04159}, 2020.

\end{thebibliography}
}

\end{document}